%% file: zimmerman2026isrrarxiv.tex
\documentclass{styles/svproc}
\usepackage{url}

\usepackage{amsmath,amsfonts}
\usepackage{graphics}           
\usepackage{amssymb}            
\usepackage{graphicx}
\usepackage{algorithm}
\usepackage[noend]{algpseudocode}
\usepackage{booktabs}
\usepackage{color}
\usepackage{pifont}
\usepackage{siunitx}
\usepackage{xcolor}
\usepackage{hyperref}

\colorlet{punct}{red!60!black}
\definecolor{background}{HTML}{EEEEEE}
\definecolor{delim}{RGB}{20,105,176}
\colorlet{numb}{magenta!60!black}

\input{macros}

\lstdefinelanguage{json}{
    basicstyle=\normalfont\ttfamily,
    showstringspaces=false,
    breaklines=true,
    frame=lines,
    backgroundcolor=\color{background},
    literate=
     *{0}{{{\color{numb}0}}}{1}
      {1}{{{\color{numb}1}}}{1}
      {2}{{{\color{numb}2}}}{1}
      {3}{{{\color{numb}3}}}{1}
      {4}{{{\color{numb}4}}}{1}
      {5}{{{\color{numb}5}}}{1}
      {6}{{{\color{numb}6}}}{1}
      {7}{{{\color{numb}7}}}{1}
      {8}{{{\color{numb}8}}}{1}
      {9}{{{\color{numb}9}}}{1}
      {:}{{{\color{punct}{:}}}}{1}
      {,}{{{\color{punct}{,}}}}{1}
      {\{}{{{\color{delim}{\{}}}}{1}
      {\}}{{{\color{delim}{\}}}}}{1}
      {[}{{{\color{delim}{[}}}}{1}
      {]}{{{\color{delim}{]}}}}{1},
}

\begin{document}
\mainmatter              
\title{SignScene: Visual Sign Grounding \\ for Mapless Navigation}
\titlerunning{SignScene}  
%

\author{Nicky Zimmerman$^{1}$, Joel Loo$^{1}$, Benjamin Koh$^{2}$, Zishuo Wang$^{1}$ and David Hsu$^{1}$}


%
\authorrunning{Nicky Zimmerman et al.} 
%
%
\institute{Smart Systems Institute, National University of Singapore, 3 Research Link,  117602, Singapore.\\
\email{nicky.zi@nus.edu.sg},
\and
School of Computing, National University of Singapore, 13 Computing Drive, 117417, Singapore.}


\maketitle              

\begin{abstract}
Navigational signs enable humans to navigate unfamiliar environments without maps. This work studies how robots can similarly exploit signs for mapless navigation in the open world. A central challenge lies in interpreting signs: real-world signs are diverse and complex, and their abstract semantic contents need to be grounded in the local 3D scene. We formalize this as \textit{sign grounding}, the problem of mapping semantic instructions on signs to corresponding scene elements and navigational actions. Recent Vision-Language Models (VLMs) offer the semantic common-sense and reasoning capabilities required for this task, but are sensitive to how spatial information is represented. We propose \method{}, a sign-centric spatial–semantic representation that captures navigation-relevant scene elements and sign information, and presents them to VLMs in a form conducive to effective reasoning. We evaluate our grounding approach on a dataset of 114 queries collected across nine diverse environment types, achieving 88\% grounding accuracy and significantly outperforming baselines. Finally, we demonstrate that it enables real-world mapless navigation on a Spot robot using only signs.
\end{abstract}

\begin{figure*}[t]
  \centering
  \includegraphics[width=0.99\linewidth]{pics/Motivation.pdf}
  \caption{We present \method{}, 
  our proposed sign-centric spatial representation for grounding signs.
     We highlight the challenges of sign grounding: (i) common sense - the sign elements can refer to the semantic notion of a straight-ahead path, rather than a geometric feature, (ii) complex instructions - signs include multiple symbols that correspond to navigational actions and need to be understood as a sequence of actions (iii) partial observability - signs may refer to elements that are not visible when viewing the sign.}
   \vspace{-10pt}
  \label{fig:motivation} 
\end{figure*}

\section{Introduction} \label{sec:intro}
Navigational signs are ubiquitous wayfinding aids in human environments which abstractly describe local actions for reaching goals---\eg{} indicating which path to take at a junction, or which exit to follow. Humans routinely use signs to make sequences of local decisions to reach distant goals without a global map. However, many robot systems rely primarily on prebuilt maps to plan and execute long-range paths to reach faraway locations. Enabling robots to interpret and act on signs offers a promising alternative paradigm for mapless navigation in previously unseen, human-built environments. We define mapless navigation as navigation without prebuilt maps, such as sensor-based global maps or public maps (i.e. floor plans).



However, interpreting navigational signs into actionable robot behaviors remains a significant challenge. Such signs associate locations with navigational instructions using natural language or abstract symbols. To use them, a robot must parse and ground their semantic content---\ie{}, map abstract directions and referenced locations to concrete actions in a rich, 3D environment. This is difficult for three reasons. 

\begin{enumerate}
 \item  \textbf{Common Sense:} navigational instructions encode abstract, human-centric concepts that demand common-sense and contextual understanding: \eg{} a symbol to indicate ``take the escalator'' or an arrow to represent ``continue straight ahead'' when none of the feasible paths are strictly straight(\figref{fig:motivation}). 
    \item \textbf{Complex instructions:} open-world signs are challenging to parse due to diversity and complexity in symbols~(\figref{fig:motivation}), layouts, and notational conventions. In more extreme cases, directions may be expressed compositionally through multiple icons, requiring reasoning to interpret. 
    \item  \textbf{Partial observability:} because signs provide only an abstracted and partial representation of the scene, grounding them requires a coherent spatial model of the local environment. For example, the escalator is not visible in \figref{fig:motivation}.
\end{enumerate}
Overall, prior approaches fail to fully address these challenges, often interpreting directions with simple geometric heuristics that ignore semantic context~\cite{liang2020wrss} or struggling with the complexity of real-world signage~\cite{agrawal2025arxiv,chandaka2025corl}.
A single image containing a navigational sign may have a partial view of the scene, as indicated by the third challenge, thus making a direct query with a VLM insufficient. On the other hand, geometry-based modular approaches are likely to fail when the navigational instruction is ambiguous, as suggested by the first challenge. 

Our main contribution is a system that addresses the challenge of parsing and grounding signs across diverse, complex scenes. To do so, we define the \textit{sign grounding problem} as determining the appropriate local action required to reach a location referenced in a sign, by grounding its abstract navigational direction to a concrete path in the scene. To construct a coherent spatial representation of the scene, we assume access to RGB observations and robot odometry. 

Our key insight is that the semantic common-sense and reasoning required for this task can be effectively addressed with Vision-Language Models (VLMs). To leverage VLMs, spatial and semantic scene information need to be conveyed in a representation aligned with the visual modality of VLMs. Additionally, constructing a sufficient spatial-semantic representation requires exploration, to capture all the regions and navigational structures that are associated with the navigational cues in the sign.

To this end, we propose \method{}, an abstract metric–semantic representation that organizes navigation-relevant scene information in a frame aligned with a sign’s location and orientation, and renders it as a simplified 2D diagram. This representation enables VLMs to reason effectively about spatial relationships and to interpret navigational instructions as intended by human designers. 
Additionally, we introduce an active perception system for parsing and grounding signs that constructs \method{} through exploration and active mapping. Suitable for real-world deployment, our system selects goal-relevant signs and explores the local scene to construct \method{}, enabling mapless navigation using signs. While a full navigation pipeline is out of scope for this paper, we plan to extend \method{} in future work.

Our approach addresses the challenge of open world signage using foundation models. We leverage a VLM to extract navigational symbols from signs and then associate them to objects in the scene via an open world object detector. To ground abstract, human-centric concepts that are potentially geometrically ambiguous, we construct VLM-aligned representation for grounding. Lastly, we mitigate partial observability by exploring and exploiting a local environment around the sign, instead of a single egocentric view.


Our experimental results demonstrate that \method{} enables VLMs to effectively parse and ground navigational signs, achieving 88\% accuracy across diverse environments and substantially outperforming recent baselines such as ReasonNav~\cite{chandaka2025corl}. We find that key design principles of \method{}---providing a simplified scene representation that captures task-relevant elements---are critical to performance. For this evaluation, we created a benchmark comprising trajectories in which an agent observes a sign and explores its surrounding area. The benchmark spans nine different environment types, including hospitals, malls, campuses, train stations, and urban outdoor settings. Finally, we demonstrate real-world mapless navigation by deploying \method{} on a Spot robot: when initialized outside an unseen building, the robot successfully detects and interprets signs, grounding their instructions into concrete subgoals that guide it toward a specified destination.
Therefore, our key claims are that (i) \method{} effectively parses and grounds navigational signs on our offline dataset (ii) our active perception pipeline can autonomously acquire a \method{} in an online fashion. These claims are backed up by the paper and our
experimental evaluation.

\section{Related Work} \label{sec:related}
\subsection{Navigational Signs}
Agrawal \etalcite{agrawal2025arxiv} formalized the problem of navigational sign understanding and proposed a VLM-based approach for parsing complex signs that include both symbols and text, coupled with non-trivial directions. 
Liang~\etalcite{liang2020iros,liang2020wrss} extracted text and direction information from signs, using them to guide local mapless navigation.
Chandaka~\etalcite{chandaka2025corl} briefly addressed utilizing signs for navigation. However, both approaches target very simple signs with 4 cardinal directions. 
Zimmerman~\etalcite{zimmerman2025ral} demonstrated that they can handle more complicated navigational signs, but their work focused on localization in public maps. Chen~\etalcite{chen2025ral} utilized signage for exploration but they only parsed locational signs such as shop names, with no directional instructions. 

\subsection{Spatial Representation for Visual Navigation}
Navigation was traditionally implemented using dense geometric representations~\cite{macenski2023ras}, such as occupancy grid maps or voxel grids. In the past few years, with the rise of LLMs and VLMs, research has seen a surge of interest in vision-language navigation (VLN)~\cite{wu2024nca}. Recent works incorporate semantic scene understanding into these dense representations~\cite{alama2025arxiv} that can be used for VLN~\cite{kim2025arxiv}. However, these approaches have a large memory footprint and the integration of navigational signs is not trivial.

Topological mapping is another common spatial representation for navigation. Object goal navigation tasks were implemented using scene graphs~\cite{loo2025ijrr,maggio2024ral}. And while LLMs showed good performance locating the desired object in the scene graph's structure, their ability to infer relative directions in this format is weak. Therefore, this representation is not suitable when grounding navigational instructions that capture information about relative directions. 

While Li \etalcite{li2024arxiv} conclude that modern VLMs struggle to reason on topview maps, this representation is increasingly common for robot navigation. 
Loo \etalcite{loo2024icra} utilize pre-built top-view maps to generate a 2D topological representation for navigation based on action primitives, and Gao \etalcite{gao2024intentionnet} demonstrated kilometer-scale navigation with top-down publicly available maps. Our approach focuses on local planning, and does not require a pre-built map.

Zhong \etalcite{zhong2024arxiv} used semantically-rich top-view map for zero-shot object-goal navigation. Zhang \etalcite{zhang2025arxiv} presented their top-down Annotated Semantic Map as a memory efficient representation for robot navigation. Chandaka \etalcite{chandaka2025corl} also utilize a semantic top-down map for navigation in human-oriented environments. We also adopt a 2D abstract map as a representation for spatial reasoning, but unlike the above works we do not limit ourselves to indoor environments, and we put an emphasis on navigational signs and their induced behaviors. 


While VLN~\cite{park2023air,krantz2020beyond} have shown impressive results without an explicit spatial representation, it requires an oracle or a human to provide step-by-step navigational instructions. Our approach grounds navigational signs in the robot's local environment, to generate explicit navigational instructions. \method{} can be complementary to VLNs, if the instructions are generated as natural language.



\begin{figure*}[t]
  \centering
  \includegraphics[width=0.98\linewidth]{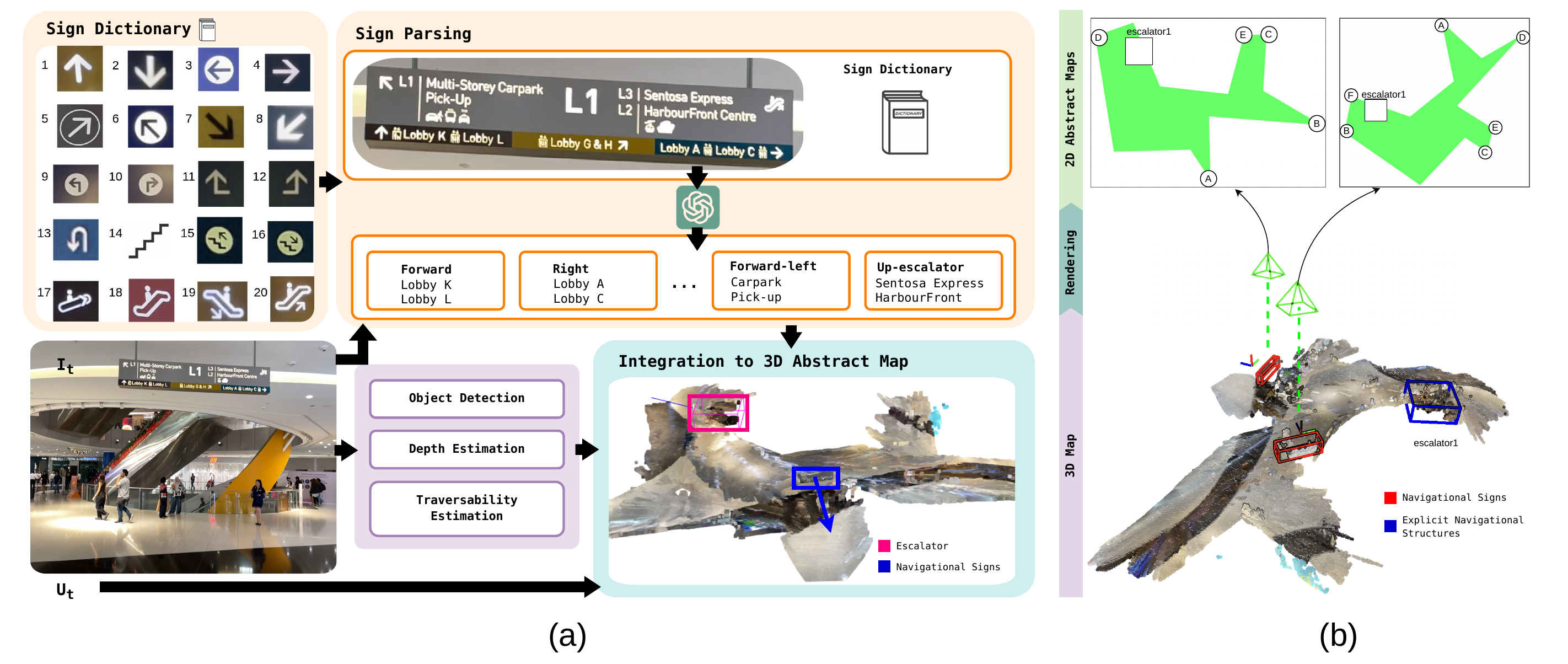}
  \caption{(a) \textbf{\method{} Construction.} Given an RGB stream $I_t$ and odometry input $U_t$, navigation-relevant objects such as escalators and signs are detected by an open world object detector, and depth is inferred with a monocular depth estimator. The sign parsing module extracts the spatial symbolic information in the detected signs. The traversability model is used as a proxy for topology. 
  In the sign parsing module, we use in-context learning to improve the sign understanding performance. We include in the prompt a sign dictionary, with examples of commonly encountered signs and their labels. The parsing output associates each location with a navigational instruction.
  (b)~\textbf{\method{}.} The 3D map represents signs and explicit navigational structures as abstract objects, and implicit paths as dense point clouds, and can be rendered into top-down, sign-centric views for VLM reasoning. In the rendered \method{}, letters refer to frontiers and objects are represented as boxes. \method{} can store more than one sign, and the rendered 2D abstract map would be rotated to align with the specific sign's cardinal axes. }
 
  \label{fig:atom} 
  \vspace{-0.2in}
\end{figure*}

\section{Problem Formulation}

To generate goal-reaching actions based on navigational signs, two key perceptual problems need to be addressed: (i) \textbf{sign understanding} to parse a sign's semantic content; (ii) \textbf{sign grounding} to ground instructions encoded in signs to the local scene. We represent a navigational sign $\navsign$ as a tuple $(b,p,\navcues)$. $b$ is the 3D oriented bounding box enclosing the sign. $p \in \mathbb{R}^6$ is its 6-DoF pose, defined relative to $b$ with the $x$-axis pointing toward the sign's front face and the $y$- and $z$-axes aligned with the $b$'s edges. $\navcues$ encodes the semantic content of the sign as a collection of tuples $\{(\signloc[1], \signdir[1]), \cdots, (\signloc[J], \signdir[J]) \}$, where each tuple associates a location $\signloc[j]$ with a navigational instruction $\signdir[j]$. 
A navigational instruction $\signdir$ identifies a spatial region $\dirregion\subseteq\mathbb{R}^3$ in the local scene possessing navigational affordances through which $\signloc$ may be reached. Instructions $\signdir$ typically take two forms: directional arrows, which point toward spatial regions like a path entrance or an access point~\cite{rooke2023wayfinding}; and object symbols (\eg{} escalator, staircase, or lift) which are semantic labels directly specifying a spatial region $\dirregion$.

\textbf{Sign understanding} is the problem of parsing a single RGB image of a navigational sign into the set $\navcues = \{(\signloc[1], \signdir[1]), \cdots, (\signloc[J], \signdir[J]) \}$ expressed in the sign. This formulation generalizes the definition in Agrawal~\etalcite{agrawal2025arxiv}, which only considers instructions $\signdir$ in the form of directional arrows. 

\textbf{Sign grounding} is a perceptual grounding task: it maps an instruction $\signdir[j]$ from a given sign $\navsign$ to a point in its corresponding region, $p\in\dirregion[j]$. This can serve downstream tasks like sign-based robot navigation, where instructions need to be grounded into concrete subgoals for navigation. We assume RGB-D images and corresponding poses as inputs, enabling consistent 3D spatial and semantic understanding of the local scene for grounding.

\section{\method} \label{sec:approach}

We introduce \method{}, an abstract metric-semantic representation that captures a navigational sign together with the scene information needed to interpret its content. \method{} is built from observations gathered in the sign's vicinity and is then used to reason about the path toward a given goal $g$, when $g$ is referenced by the sign. \secref{sec:representation} introduces the structure of \method{}. The following sections describe how sign and scene information are integrated into it: \secref{sec:signunderstanding} covers the extraction of a sign's semantic content, and \secref{sec:mapbuilding} describes how RGB observations and robot odometry are fused into a representation of the local scene's geometry and objects. \secref{sec:active} then presents an exploration strategy that actively gathers the observations needed to construct \method{} on a real robot. Finally, \secref{sec:spatialreasoning} describes how a VLM is queried with \method{} to perform sign grounding. Concretely, it returns an $\mathrm{SE}(2)$ subgoal that realizes the navigational instruction $\signdir$ for reaching the specified goal $g$.



\subsection{\method{} Structure} \label{sec:representation}

\method{} is a spatio-semantic representation of the local scene designed to enable sign grounding with VLMs by providing a simplified view for decision-making. Formally, \method{} is a tuple $(S, R_{\text{obj}}, R_{\text{dir}})$, where $S = \{\navsign_1, \dots, \navsign_n\}$ is the set of signs in the local scene, and $R_{\text{obj}}$ and $R_{\text{dir}}$ are the spatial regions that the signs may refer to, through object symbols or directional arrows respectively. Specifically, $R_{\text{obj}}$ is the set of navigational structures whose semantic labels match symbol classes (\eg{}, escalator, lift, staircase), while $R_{\text{dir}}$ is a more open-ended set of traversable regions such as corridors, paths, and access points that arrows may point toward. Together, $R_{\text{obj}}$ and $R_{\text{dir}}$ are intended to cover all regions in the vicinity of the sign that participate in the local topological connectivity, giving the VLM the spatial context needed to ground each sign's instructions. \method{} renders these elements in a simplified 2D top-down view for VLMs' consumption.

\textbf{2D abstract top-view map ($\mapsimple$).} $\mapsimple$ is the simplified 2D representation generated for VLMs, and follows three design principles. First, $\mapsimple$ is a top-view projection of the scene, preserving the layout and relative positioning of spatial regions needed to ground a sign's instructions. Second, $\mapsimple$ is centred on, and oriented according to the sign's pose $p$, with ``up'' aligned with the $x$-axis of $p$'s frame, to preserve the reader's viewpoint and ensure instructions are interpreted as intended. Third, $\mapsimple$ minimises visual distractors to support effective spatial reasoning---navigational structures are rendered as labelled 2D bounding boxes and traversable areas as a shaded polygon, a simple representation readily understood by VLMs. Given a sign $\navsign_i$, $\mapsimple$ is rendered from an underlying 3D representation of the local scene, $\mapfull$.

\textbf{3D map ($\mapfull$).} $\mapfull$ is a geometrically consistent 3D representation of the local scene, constructed by integrating RGB-D observations and odometry. It instantiates the elements of \method{} in 3D: each sign $\navsign \in S$ is realized by its bounding box and pose; each region in $R_{\text{dir}}$ is represented as a point cloud capturing a traversable area such as a corridor, path, or access point; and each navigational structure in $R_{\text{obj}}$ is represented as a labeled point cloud with an associated 3D bounding box. Since regions can be open-ended, $\mapfull$ encodes all traversable areas in the local scene, providing a superset from which the regions referred to by directional arrows are drawn.

\textbf{\method{} rendering.} To obtain $\mapsimple$, $\mapfull$ is rigidly transformed into the sign's coordinate frame, $p$, and projected into a top-down view~(\figref{fig:atom}(b)). Traversable areas are projected onto a 2D grid, from which the boundary contour is extracted as a simplified polygon. Navigational structures' estimated oriented bounding boxes are projected as oriented 2D rectangles, annotated with their semantic class and instance identifier. To cast sign grounding as a discriminative multiple-choice problem amenable to VLM reasoning, candidate grounding regions are identified and labeled with alphabetical identifiers. Navigational structures form one set of candidates. For traversable areas, candidate points are sampled at convex protrusions of the contour polygon (one point per piecewise-convex segment), approximating the discrete paths a directional instruction may refer to.

\subsection{Sign Understanding} \label{sec:signunderstanding}

We build on the sign understanding approach of \cite{agrawal2025arxiv}, querying a VLM to parse an image of a sign into navigational cues $\navcues$. This leverages the VLM’s rich understanding of text and common human symbols, and its strong capability to reason about complex, compositional instructions conveyed through text and icons. To improve robustness in parsing symbolic content, we prompt the VLM with a small set of symbol prototypes collected from diverse environments (\figref{fig:atom}(a)). This in-context learning strategy enables the VLM to generalize beyond cardinal directions and handle a broader range of navigational structures, such as stairs and escalators. 
Concretely, our sign understanding module prompts the VLM to extract the locations and associate each location with an element from the set of navigational instructions $\signdir[j]$, given by the sign dictionary.
The sign understanding module assumes that the camera is aligned with a sign-centric frame, \ie{}, viewing the sign head-on. For real-world deployment, we integrate a sign alignment module (\secref{sec:active}) that actively aligns the robot with a detected sign to satisfy this assumption.  We ensure that the image crops used for the visual symbol prototypes in the dictionary do not overlap with any testing data.

\begin{figure*}[t]
  \centering
  \includegraphics[width=0.98\linewidth]{pics/VisualGrounding.pdf}
  \caption{\textbf{Reasoning with \method{}}. Given a desired goal (``Terrace''), we first search the best matching location stored in \method{}'s signs, based on a text similarity score. For the selected sign, the 3D map is rotated such that the sign is aligned with the canonical axes. Then we render \method{} to a simplified 2D map, annotated with objects as boxes and frontiers as letters, which the VLM can reason on, and ground the navigational instruction associated with the desired goal. The visual grounding outputs a 3D goal in the robot's local coordinate system. \textbf{Benchmark Format}. The visual grounding problem is phrased as a multiple choice question on an image, with 4 answers annotated as (1,..,4), each represents a potential path, $l_i$. We use the predicted 3D goal to choose the closest answer that represents the correct path. 
  }
  \label{fig:reasoning} 
  \vspace{-0.2in}
\end{figure*} 

\begin{figure*}[t]
  \centering
  \includegraphics[width=0.98\linewidth]{pics/ActiveEngine2.pdf}
  \caption{ \textbf{Real-world Deployment.} 
  First, the active perception pipeline actively detects navigational signs with open world object detector and visually servoes to a candidate sign to better parse it. The active \method{} system then actuates the Spot robot arm to tackle signs of varying heights.
  Once the navigational instructions are extracted, the pipeline enters its active mapping phase, where it constructs a \method{} representation by locally exploring the scene. Lastly, it grounds the navigational instruction associated with the desired goal, by reasoning with \method{}, as explained in \figref{fig:reasoning}.}
  \label{fig:signscene} 
    \vspace{-0.2in}

\end{figure*} 

\subsection{Constructing \method{}} \label{sec:mapbuilding}
$\mapfull$ is built online from incoming RGB observations, $\obs$, and odometry, $\odom$~(\figref{fig:atom}(a)). 3D geometry is estimated by applying monocular depth estimation to RGB inputs. We use these inputs to infer the key scene elements in $\mapfull$, \ie{} navigational signs, navigational structures and traversable area.

\textbf{Navigational signs.} We detect, segment, and temporally fuse navigational signs across an input RGB-D sequence to estimate each sign’s 3D location and orientation. For signs observed from a viewpoint suitable for parsing with VLMs, we extract their semantic content $\navcues$ and associate it with the corresponding fused sign instance. A viewpoint is considered suitable if the sign is within a distance threshold $\tau_{\text{dist}}$ of the robot and its surface normal $n_s$ is aligned with the robot’s viewing direction within an angular threshold $\tau_{\text{angle}}$. We empirically determine these thresholds through tests with VLMs including GPT-5 and Gemini-2.5-Pro.

Sign detection and segmentation are performed independently for each image $I_t$. We prompt an open-set object detector with the query ``signs'' to obtain bounding boxes, which are used to prompt SAM2~\cite{ravi2025iclr} for masks $\signmask$. Using the corresponding depth image $\depth$, we extract 3D point cloud segments $\depthseg$ for each detected sign using $\signmask$, estimate their centroids and surface normals, and cluster detections across time based on similarity in centroids and normals. Normals are estimated using the plane fitting method of \cite{zhou2018arxiv}.

Detected signs are parsed using the sign understanding module (\secref{sec:signunderstanding}). For each sign cluster $i$, we aggregate the navigational cues $\navcues^i_{t_{\text{start}}}, \dots, \navcues^i_{t_{\text{end}}}$ extracted across the sequence. The final cue set $\navcues^i_{\text{merged}}$ is produced by selecting the most frequently predicted navigational instruction for each referenced location, and is then associated with the fused sign instance.

\textbf{Traversable areas.} These regions correspond to areas of the scene designed to facilitate navigation. We extract path masks $\pathmask$ from each image $\obs$ with a path segmentation model---we use GeNIE~\cite{wang2025arxiv}, a traversability model fine-tuned on human-annotated data to segment paths preferred by humans. We use $\pathmask$ to extract path point cloud segments from each image $\obs$ and  fuse them to obtain a consolidated 3D pointcloud representation of navigable paths in the local scene.

\textbf{Explicit navigational structures.} These are semantically meaningful scene elements commonly referenced in navigational signage and well supported by object detection models. We detect these structures in each image $I_t$ using an open-set object detector prompted with a small, fixed vocabulary of navigation-relevant classes drawn from the sign dictionary (\figref{fig:atom}(a)). Although we use a fixed prompt in this work, our approach naturally extends to larger or open sets of navigational structures. For each detection, we use the predicted bounding box to extract the corresponding 3D point cloud segment $\depthseg$. 
Detected structures are then temporally fused across frames using a procedure similar to \cite{zimmerman2023iros}, where detections in frame $I_t$ are matched to existing map objects $\mathbb{O}$ via the Hungarian algorithm. From the fused point cloud, we estimate an oriented 3D bounding box and encode each structure in \method{} with its 3D geometry, semantic class label, and a confidence score obtained by averaging the GroundingDINO detection scores.

\subsection{Active Perception for \method{}} \label{sec:active}

We propose an active perception system that builds a complete local \method{} representation around detected signs for reasoning about signs and  enabling sign-based mapless navigation. Properly grounding instructions from signs requires sufficiently complete knowledge of the local scene's geometry and navigational structures. However, occlusions and partial observability from a viewpoint facing the sign often prevent the robot from perceiving all relevant elements. Our active perception system~(\figref{fig:signscene}) plays a critical role in ensuring a sufficiently complete \method{} is constructed.

Our active perception system uses an open-set object detector to detect navigational signs. Upon detection, our system servos the robot to face the sign head-on, using estimates of the sign’s 6-DoF pose computed from depth. VLMs' ability to parse signs noticeably degrades at oblique viewing angles. Since signs may appear at varying heights, this step is critical for effective parsing. For robots with articulated sensors, \eg{} a Spot robot with an arm-mounted camera, the system further actuates the sensor to refine the viewpoint and maximize frontal alignment. The sign is then parsed (\secref{sec:signunderstanding}) and if the specified goal is indicated on the sign, the system transitions to exploration, aiming to ground the sign’s instructions within a locally constructed \method{}.

The exploration module assumes the robot starts approximately aligned with and facing the sign, providing a good initialization of \method{}'s sign-centric coordinate frame. First, the robot performs an in-place rotation at its initial pose to observe its immediate surroundings and identify candidate frontier points. Candidate frontiers are selected based on proximity to potential paths, which are inferred from the geometry of the ground-region segmentation mask introduced in \secref{sec:mapbuilding}: in a top-down view, paths typically appear as convex protrusions of the ground mask. We therefore simplify the 2D ground-region mask into a polygon and identify frontier points at the centers of these convex protrusions, treating them as candidate path entrances. In the second stage, the robot sequentially visits these frontiers to incrementally complete the \method{}, ensuring sufficient coverage of the sign, implicit paths, and explicit navigational structures for effective downstream reasoning.


\subsection{Reasoning with \method{}} \label{sec:spatialreasoning}

Given a constructed \method{}, we can perform sign grounding with VLMs and obtain a subgoal that takes the robot toward a given goal location $g$ (\figref{fig:reasoning}), if there exists a location corresponding to $g$ in the sign, \ie{} $\signloc[g]$.

\textbf{Location matching.} As a \method{} may contain more than one sign relevant to the queried goal, we match each query against all signs in the \method{}. For each sign $s^i$, we match the queried location $l_g$ against the parsed locations $l_p^1...l_p^k$ using fuzzy matching score based on Levenshtein distance~\cite{Levenshtein1965spd}, and pick the highest scoring match $l_p^j$ with score $s_p^j$. As described in \secref{sec:representation}, we render \method{} into $\mapsimple$, centred on the sign containing $l_p^j$.

\textbf{VLM grounding.} We then query a VLM to ground $l_g$. The inputs are matched location $l_p^j$ and its associated instruction $d_p^j$, and the  rendered $\mapsimple$. The VLM is instructed to select an implicit path frontier or an explicit navigational structure that is the closest to the location $l_p^j$, and aligned with $d_p^j$. The selected scene element in $\mapsimple$ is then converted to a 3D location, $\text{goal}^i$, which is associated with the matching score $s_p^j$. If a queried location is well-matched in multiple signs, we will choose $\text{goal}^i$ with the highest score $s_p^j$. This $\text{goal}^i$ can be used as the next subgoal en route to $l_g$.

For the benchmark evaluation we frame the visual sign grounding problem  as a multiple-choice question where path options are provided~(\figref{fig:reasoning}), for a queried location $l_g$. For the grounding answers $a_q^1...a_q^4$, and the timestamp $t$, we retrieve the robot's estimated pose $x_t$ and depth map $\mathcal{M}^t_{\text{depth}}$ and compute the 3D locations of the grounding answers. The selected grounding answer $a_q^{\text{selected}}$ corresponds to the 3D location that has the smallest Euclidean distance to $\text{goal}^i$.

\section{Experiments}
\subsection{Experimental Setup} \label{sec:setup}

\textbf{Benchmark and offline data collection.} To evaluate the performance of \method{} we collected data with a hand-held device. We recorded 36 sequences that contain high-resolution RGB images, sparse depth images and visual-inertial odometry. The recordings were taken in 9 different environments, including hospitals, malls, MRT stations, airport, university buildings and outdoor areas. Each recording includes at least one navigational sign, and the local environment around it. For each sequence, we compiled a list of multiple-choice visual grounding queries with annotated ground-truth for ease of evaluation. The grounding query targets a location that is included in one of the visible signs, and includes four different grounding options are annotated on a frame~(\figref{fig:reasoning}) from the sequence, with one of them representing the correct instruction grounding. Examples for the benchmark's environments and queries are shown in the video.

\textbf{Online robot system.} We deploy \method{}~(\figref{fig:signscene}) on a Spot robot with arm for real-world navigation using signs. As described in \secref{sec:active}, we utilize our active perception system as an active perception engine to construct \method{}, starting from active sign understanding and following it up by local exploration. 
The full system runs onboard a Jetson Orin AGX, with only VLM queries executed off-board.

\textbf{Models.} Across all modules, we use GroundingDINO~\cite{liu2024eccv} as our open-set object detector, GeNIE~\cite{wang2025arxiv} for path segmentation, and Metric3Dv2~\cite{hu2024tpami} for estimating monocular depth from RGB. We employ GPT-5 as our main VLM, although the online system uses Gemini-2.5-Pro for sign understanding owing to its better robustness across different cameras.

\begin{table*}[t]
  \caption{Visual grounding performance for 9 different environments. }
  \footnotesize
  \resizebox{0.94\textwidth}{!}
  {
  \begin{tabular}{@{\extracolsep{4pt}}cccccccccccccccccccc@{}}
    \toprule
  & \multicolumn{6}{c}{\textbf{Campus}} & \multicolumn{3}{c}{\textbf{Hospital A}}   & \multicolumn{5}{c}{\textbf{Hospital B}}  & \multicolumn{4}{c}{\textbf{Mall A}}  \\
    \cmidrule{2-7}\cmidrule{8-10}\cmidrule{11-15} \cmidrule{16-19}
    & S1 & S2 & S3 &  S4 & S5 & S6  & S7 & S8 & S9 & S10 & S11 & S12 & S13 &  S14 & S15 & S16 &  S17 & S18  \\
      \midrule
 Ours & 3/3 & 3/3 & 3/3 & 2/2 & 2/2 & 2/3 & 2/2 & 4/4 & 3/4 & 1/3 & 4/4 & 2/4 & 4/4 & 3/3 & 
5/5 & 2/4 & 3/4 & 4/4 \\ 
ReasonNav & 0/3 & 0/3 & 2/3 & 1/2 & 0/2 & 1/3 & 0/2 & 2/4 & 2/4 & 0/3 & 1/4 & 2/4 & 2/4 & 0/3 & 1/5 & 1/4 & 0/4 & 0/4   \\
  \end{tabular}
  }



   \resizebox{\textwidth}{!}
  {
  \begin{tabular}{@{\extracolsep{4pt}}ccccccccccccccccccccc@{}}
    \toprule
   & \multicolumn{2}{c}{\textbf{Mall B}}  & \multicolumn{2}{c}{\textbf{Mall C}} & \multicolumn{5}{c}{\textbf{MRT}} & \multicolumn{4}{c}{\textbf{Airport}} & \multicolumn{5}{c}{\textbf{Outdoors}}  & \multicolumn{1}{c}{\textbf{All}} \\
    \cmidrule{2-3} \cmidrule{4-5} \cmidrule{6-10} \cmidrule{11-14} \cmidrule{15-19}   \cmidrule{20-20}
 &  S19 &  S20 & S21 & S22 & S23 & S24 & S25 & S26 &  S27 & S28 &  S29 &  S30 & S31 & S32  & S33 & S34 & S35 & S36  & \\
      \midrule
Ours & 2/2 & 2/2 & 5/5 & 4/4 & 2/3 & 4/4 & 3/3 & 2/2 & 2/2 & 1/4 & 2/2 & 2/2 & 2/2 & 3/3 & 3/3 & 2/2 & 4/4 & 4/4 & \textbf {101/114} \\
ReasonNav & 1/2 & 0/2 & 2/5 & 2/4 & 0/3 & 1/4 & 1/3 & 1/2 & 0/2 & 1/4 & 0/2 & 0/2 & 0/2 & 1/3 & 1/3 & 1/2 & 0/4 & 3/4 & 30/114  \\
  
    \bottomrule
  \end{tabular}
  }
  \label{tab:groundingresults}
  \vspace{-0.1in}
\end{table*}

\subsection{Visual Sign Grounding}

We evaluate sign grounding with \method{} on our benchmark, to support our first claim that we can parse and visually ground navigational signs, and report performance in \tabref{tab:groundingresults}. Since the evaluation is on an offline dataset,  \method{} is constructed frame-by-frame without our active perception modules, showcasing the contribution of our representation. 
We also compare against prior work using signage for navigation. In particular, we evaluate ReasonNav~\cite{chandaka2025corl}, a recent state-of-the-art method that similarly employs VLMs for sign-based reasoning and navigation. To do so, ReasonNav builds a global map using LiDAR-based SLAM~\cite{Macenski2021joss}, and embeds the detected signs in the global map, which the VLM reasons with.

We achieve a success rate of 101/114 (88.6\%), demonstrating that our proposed representation is capable of grounding navigational signs across diverse environments. 
We find that VLM reasoning with \method{} significantly outperforms ReasonNav on the sign grounding task. As ReasonNav also employs foundational models, our superior results can be attributed to our \method{} representation, rather than the power of foundation models.  We identified four key differences contributing to this disparity:

\textbf{Manhattan world assumption.} ReasonNav handles only simple navigational instructions. It considers instructions in the form of only 4 cardinal directions($\uparrow, \downarrow, \leftarrow, \rightarrow$), essentially making a Manhattan world assumption and supporting only simple environments and navigational signs. Instead of decomposing signs to navigational instructions as proposed by our approach, ReasonNav associates locations only with a subset of cardinal directions, which leads to failures in sequences with more complex scenes and navigational signs. Our approach is able to operate without the Manhattan world assumption owing to a more general sign understanding approach.
     
\textbf{Partial observability.} ReasonNav does not perform exploration or address the partial observability of the local scene when grounding signs. Specifically, it selects the next action immediately after parsing a navigational sign. In many scenes, key paths or navigational structures in the surrounding scene are obscured or can only be partially observed when viewing the sign. ReasonNav thus makes decisions based only on partial observations of the scene, contributing to high failure rates. In contrast, \method{} fuses information spatially and temporally, utilizing the full sequence of frames, to build up a more complete representation of the local scene prior to sign grounding.
     
\textbf{Topology extraction.} ReasonNav makes stronger assumptions in identifying paths in the scene compared to \method{}. ReasonNav detects paths based only on geometry, by segmenting out a flat ground plane. In contrast, \method{} considers both geometry and semantics, by using a general path segmentation model trained to capture a broader, human-defined notion of paths. This allows \method{} to capture more general paths beyond corridors, such as outdoor scenes where paths are defined by terrain types and textures (\eg{} dirt road surrounded by grass). ReasonNav cannot handle outdoor environments, and can fail to correctly identify paths in environments owing to brittleness in its path detection, leading to incorrect grounding of navigational instructions.
      
\textbf{Representation frame.} ReasonNav operates on a global map, requiring the VLM to consider the sign orientation and position when grounding the sign. In our sign-centric representation, we capture the local scene around the sign and the rendered \method{} is orientated according to the signs' cardinal axes.

\subsection{Ablations}

We ablate on our parsing module~(\secref{sec:signunderstanding}), reported in \tabref{tab:ablations}. For each of the 114 grounding questions, we test the correctness of the parsing for the location of interest. In the parsing result, if the navigational instruction associated with the location of interest matches the hand-annotated ground truth navigational instruction, then the parsing is considered successful. First, we run the evaluation for the original sign parsing method of Agrawal~\etalcite{agrawal2025arxiv}, which scored 38/114. Then we used the parsing prompt without the in-context learning example~(\figref{fig:atom}), similarly to ReasonNav's approach, scoring 53/114. Next, we enhance our parsing prompt with an in-context learning example, improving the score to 61/114. Lastly, we evaluate our parsing method which includes an additional temporal filtering, where we merge multiple observations of the same sign. With this, we achieve 93 correct parsing results out of 114, demonstrating that our approach provides improvements over previous works~\cite{agrawal2025arxiv,chandaka2025corl}. 

\begin{table}[t!]
  \caption{Navigational signs parsing performance on the dataset}
  \centering
  \footnotesize
  \resizebox{0.75\textwidth}{!}
  {
  \begin{tabular}{@{\extracolsep{4pt}}ccc@{}}
    \toprule
    Ablations &  In-context Learning & Parsing Success \\
    \midrule
     Agrawal~\etalcite{agrawal2025arxiv} & \xmark & 38/114 \\
    Chandaka~\etalcite{chandaka2025corl}& \xmark & 53/114\\
    Ours (no temporal filtering) & \cmark &61/114\\
    Ours & \cmark & 93/114\\
    \bottomrule
  \end{tabular}
  }
  \label{tab:ablations}
  \vspace{-0.2in}
\end{table}

\subsection{Real-world demonstration}

\begin{figure*}[t]
  \centering
  \includegraphics[width=0.98\linewidth]{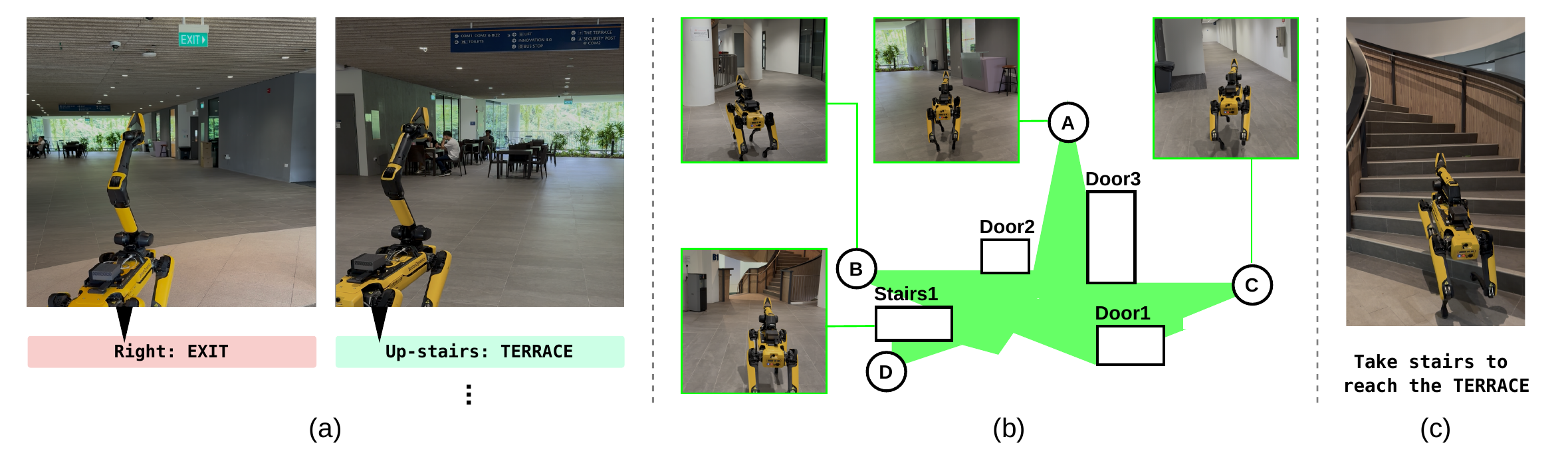}
  \caption{\textbf{Real-world mapless navigation with signs.} The robot is given ``TERRACE'' as a goal: \textbf{(a)} it explores multiple signs until it finds one with information relevant to the goal, then \textbf{(b)} explores the local environment to build \method{}, enabling successful grounding and guiding the robot to take the stairs to reach ``TERRACE'' in \textbf{(c)}.}
  \label{fig:demo} 
   \vspace{-0.2in}
\end{figure*} 
In our real-world demonstration, we deploy \method{}, and validate our second claim that our active perception pipeline can autonomously acquire a \method{} using its active perception modules.
We demonstrate that \method{} enables robotic mapless navigation in unseen environments. The robot is given a goal location and initialized in the driveway of an unfamiliar building, with no prior map, and must therefore rely entirely on signs to determine appropriate actions. \figref{fig:demo} shows an example in which the robot is tasked with reaching the goal “TERRACE.” The online and active variant of \method{} (\secref{sec:setup}) is deployed on the robot. We find that \method{} enables the robot to  detect and parse navigational signs, and reliably identify the signs that contain information relevant to the goal. Our exploration module reliably identifies and explores potential paths and navigational structures, producing a sufficiently complete \method{} representation of the local scene for accurate instruction grounding. As a result, the robot correctly identifies and navigates to the stairs required to reach the “TERRACE.” \method{} construction runs onboard in real time. Sign parsing queries Gemini-2.5-Pro (3~\unit{\second}), while sign grounding queries GPT-5 (20~\unit{\second}).  A typical active \method{} execution includes one VLM call for parsing, and one VLM call for grounding, with a local exploration that takes around 2 minutes. Therefore, the VLM query times are not very significant considering the total execution time. 
This demonstrates \method{}'s effectiveness for real-world, mapless navigation using signs.

\section{Conclusion}\label{sec:conclusion}
We present \method{}, a sign-centric representation that captures navigation-relevant scene information alongside instructions from signage, and which can be rendered into a simplified 2D representation for reasoning with VLMs. \method{} enables robots to interpret and ground instructions from signs, paving the way towards mapless navigation and wayfinding in human-oriented environments. We further introduce an active perception system that constructs a \method{} through local exploration, which can be integrated into a navigation system for signage-aware mapless navigation. Our experiments show that \method{} effectively parses and grounds navigational signs with 88\% accuracy across diverse environments, outperforming existing approaches. We couple it with our active perception \method{} pipeline to demonstrate autonomous mapless navigation with signs on a Spot robot in the real world.

While the performance of our approach is stable across various environments, there are multiple challenges that remain unsolved. The approach is limited by the capabilities of the object detector. In addition, parsing of compound signs is still out of reach and is necessary for better generalization across environments. Furthermore, mapping in very crowded areas can result in incomplete traversability maps, which are crucial to inform the VLM of the spatial topology. Moreover, the VLM parsing query is time-consuming and attempting to parse multiple signs would slow the agent down. And lastly, reasoning with the VLM can also fail in certain scenarios.

\vspace{-0.1in}

\bibliographystyle{plain_abbrv}
\bibliography{glorified,new}

\begin{figure*}[t!]
  \centering
  \includegraphics[width=0.95\linewidth]{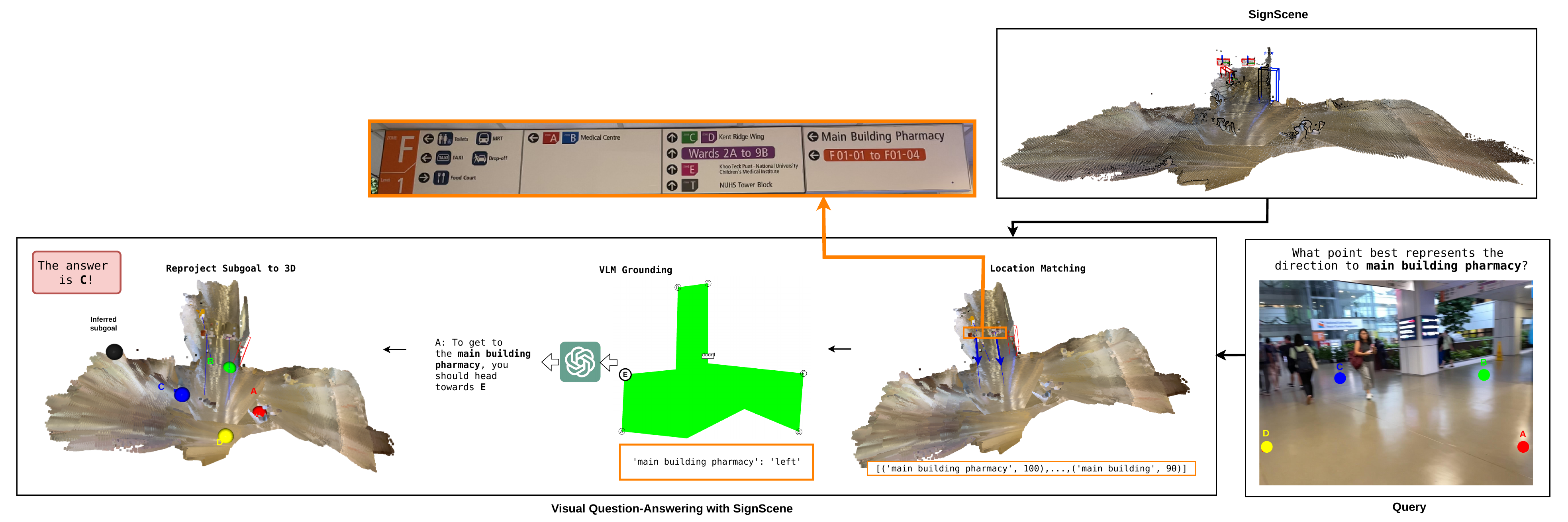}
  \caption{\textbf{Lobby of large hospital}. \method{} selects the sign in the captured by its representation (highlighted in orange) as being relevant to the query of \textbf{main building pharmacy}. It grounds the queried location to a navigation subgoal in the \method{} render. This subgoal is closest to point C in the query, resulting in the answer ``C''.}
  \label{fig:S22} 
\end{figure*} 

\section*{Appendix A: VQA Examples}

We present several VQA examples to showcase the diversity of our benchmark and to highlight \method{}'s performance across different scenes. These scenarios were collected in different environments, including malls, hospitals, subway stations, an airport and various outdoor locations. 

The selected sequences cover various topologies, such as large open spaces and complex intersections, diverging from the standard Manhattan world corridor-junction test cases. They include shiny floors, roads with markings, carpets with elaborate patterns, walking paths along grass patches - all which can challenge traversability and path segmentation models. In addition, the inclusion of glass walls and windows, flat grass patches near paths and wide open spaces present a challenge to LiDAR-based and geometry focused approaches. Another challenge comes from the dynamics - these public spaces are very crowded and rich with moving elements, from humans to strollers and baggage carts. 

\begin{figure*}[h]
  \centering
  \includegraphics[width=0.95\linewidth]{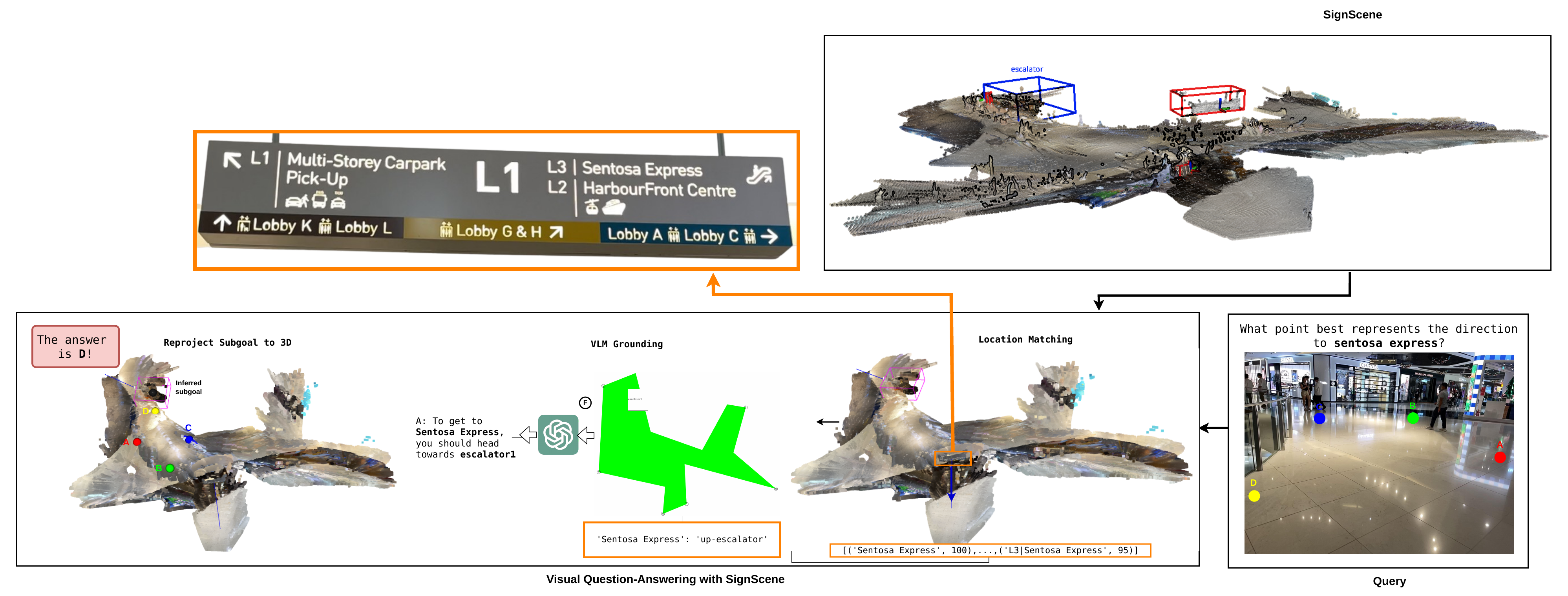}
  \caption{\textbf{Near atrium of large mall}. \method{} selects the sign in the \method{} container~(highlighted in orange) as being relevant to the query of \textbf{sentosa express}. It grounds the queried location to a navigation subgoal in the \method{} render. This subgoal is closest to point D in the query, resulting in the answer ``D''.}
  \label{fig:S26} 
\end{figure*}

We also aim to convey the variation in navigational signs' appearance, conceptual design, complexity and the open vocabulary of both text and symbols. Navigational signs also include hierarchies of spaces, such as the zone and floor as locational elements~\figref{fig:S22} with more specific locations being associated with directions.
As demonstrated by ~\figref{fig:S26}, it is not trivial to separate high-level elements that indicate the current location (L1), from similarly annotated elements that indicate a specific location is in a different level of the hierarchy, i.e. "L3 \textbar{} Sentosa Express".  

\begin{figure*}[h]
  \centering
  \includegraphics[width=0.95\linewidth]{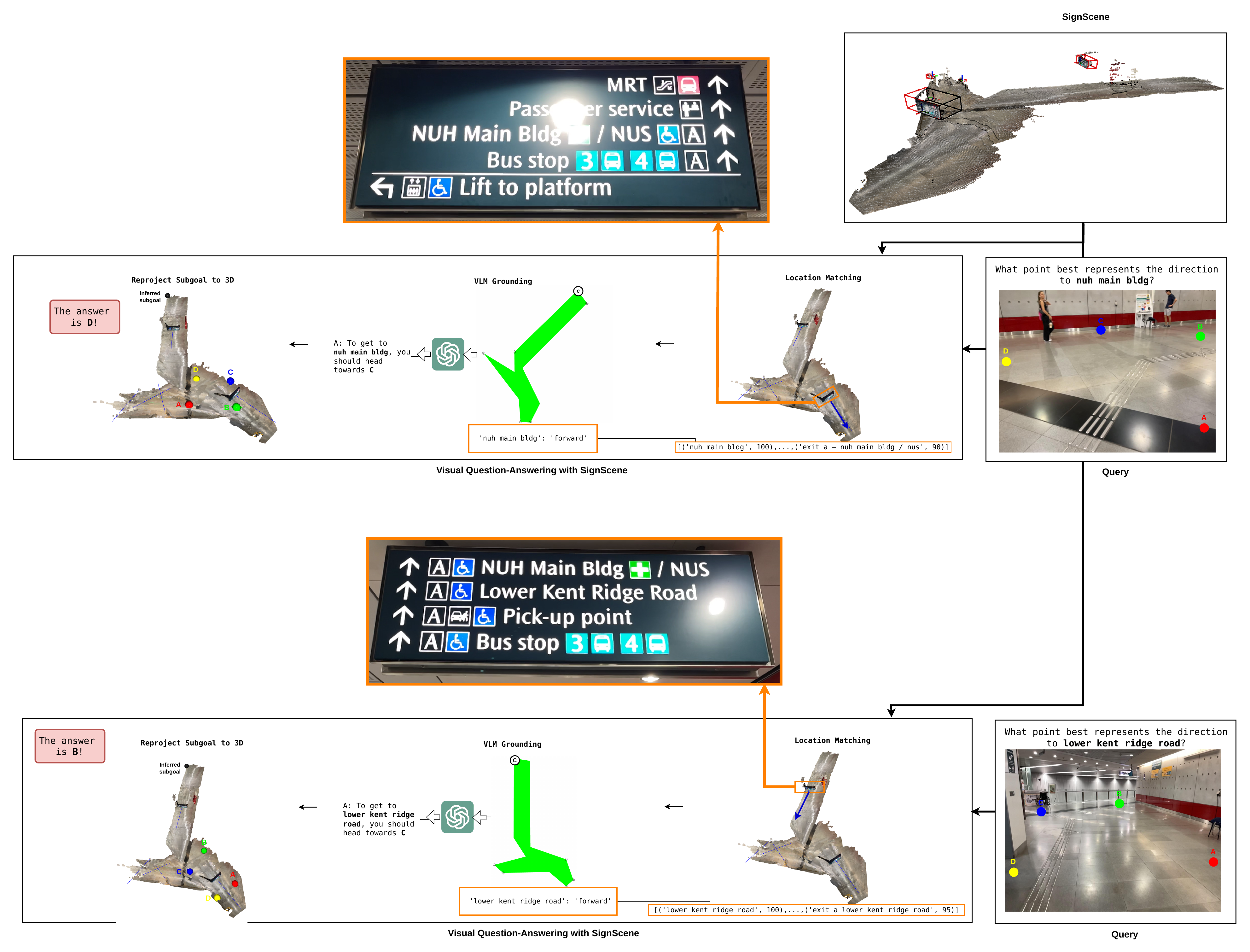}
  \caption{\textbf{Near fare gates in subway station}. We make multiple queries on a single \method{} container. For each query, \method{} selects the sign containing information relevant to the query from the \method{}. It grounds the queried location to a navigation subgoal in the \method{} render, then identifies the point in the image query closest to the computed subgoal, returning that point as the answer to the query. In (a), \method{} selects the sign highlighted in orange as being relevant to the query \textbf{nuh main building}. The computed subgoal is closest to point D in the query. In (b), \method{} selects the sign highlighted in purple as being relevant to the query \textbf{lower kent ridge road}. The computed subgoal is closest to point B in the query.}
  \label{fig:S67} 
\end{figure*}

In many of the scenarios, there was more than one navigational sign in the local environment. During the VQA, our pipeline matched the goal location with the existing signs, and centered the projected \method{} around the most relevant sign. In \figref{fig:S67} we demonstrate that \method{} can store multiple navigation signs and then ground the navigational instructions from each, by re-centering the projection around the relevant sign. We query about two locations, matched to two different navigational signs and resulting in different \method{} projections. 

\begin{figure*}[h]
  \centering
  \includegraphics[width=0.95\linewidth]{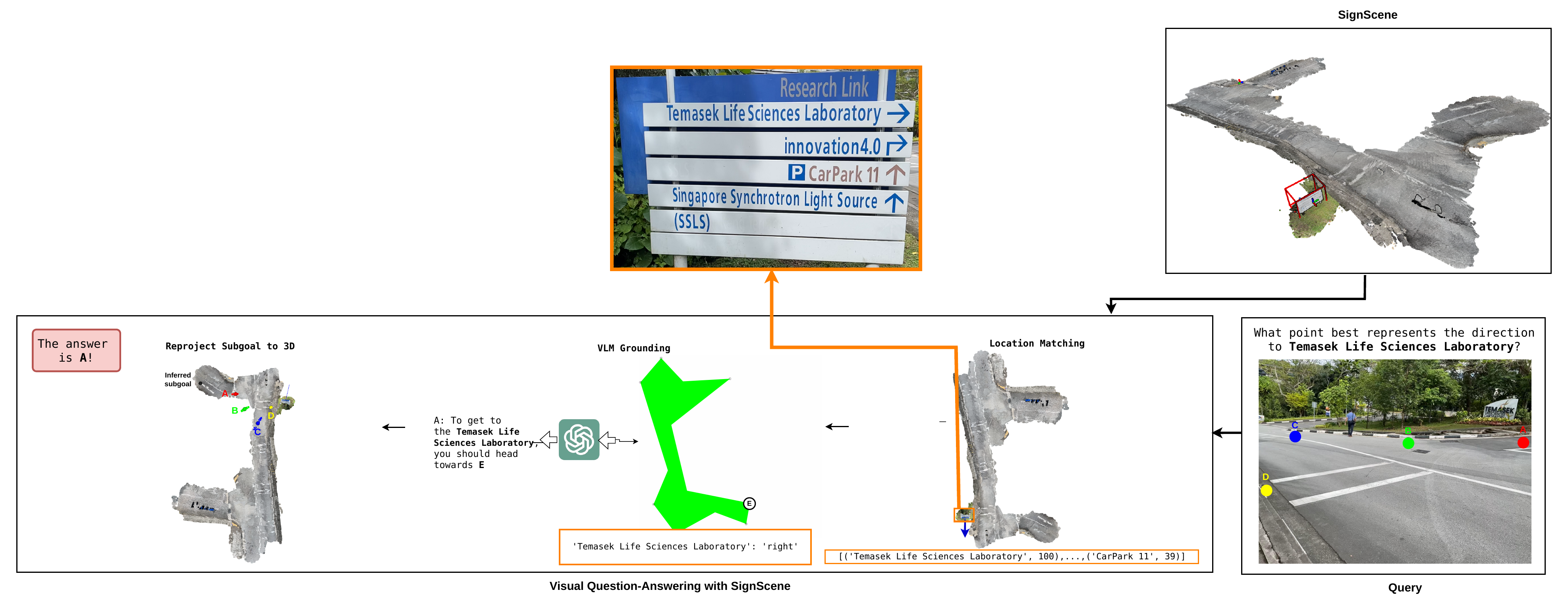}
  \caption{\textbf{Main road on university campus}. \method{} selects the sign in the \method{} container~(highlighted in orange) as being relevant to the query of \textbf{Temasek Life Sciences Laboratory}. It grounds the queried location to a navigation subgoal in the \method{} render. This subgoal is closest to point A in the query, resulting in the answer ``A''.}
  \label{fig:S40} 
\end{figure*} 

\begin{figure*}[h]
  \centering
  \includegraphics[width=0.95\linewidth]{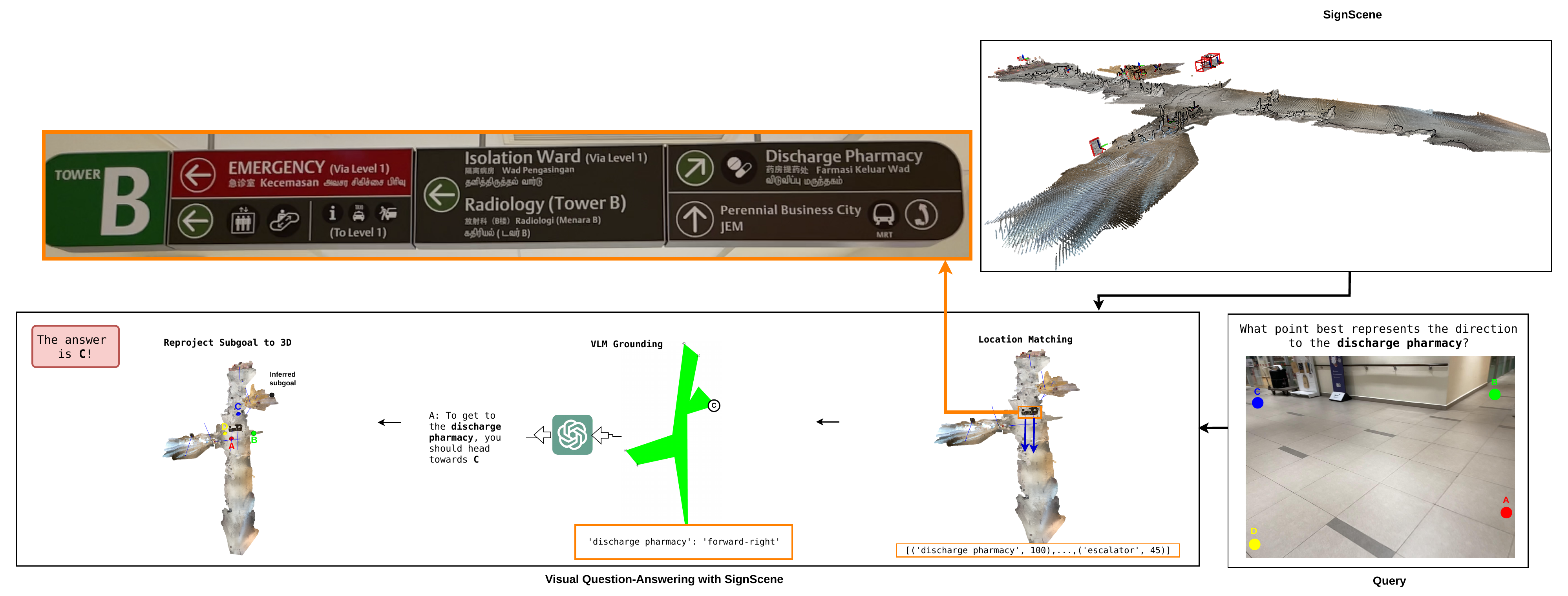}
  \caption{\textbf{Main corridor of large hospital}. \method{} selects the sign in the \method{} container~(highlighted in orange) as being relevant to the query of \textbf{discharge pharmacy}. It grounds the queried location to a navigation subgoal in the \method{} render. This subgoal is closest to point C in the query, resulting in the answer ``C''.}
  \label{fig:S48} 
\end{figure*} 

\begin{figure*}[t]
  \centering
  \includegraphics[width=0.95\linewidth]{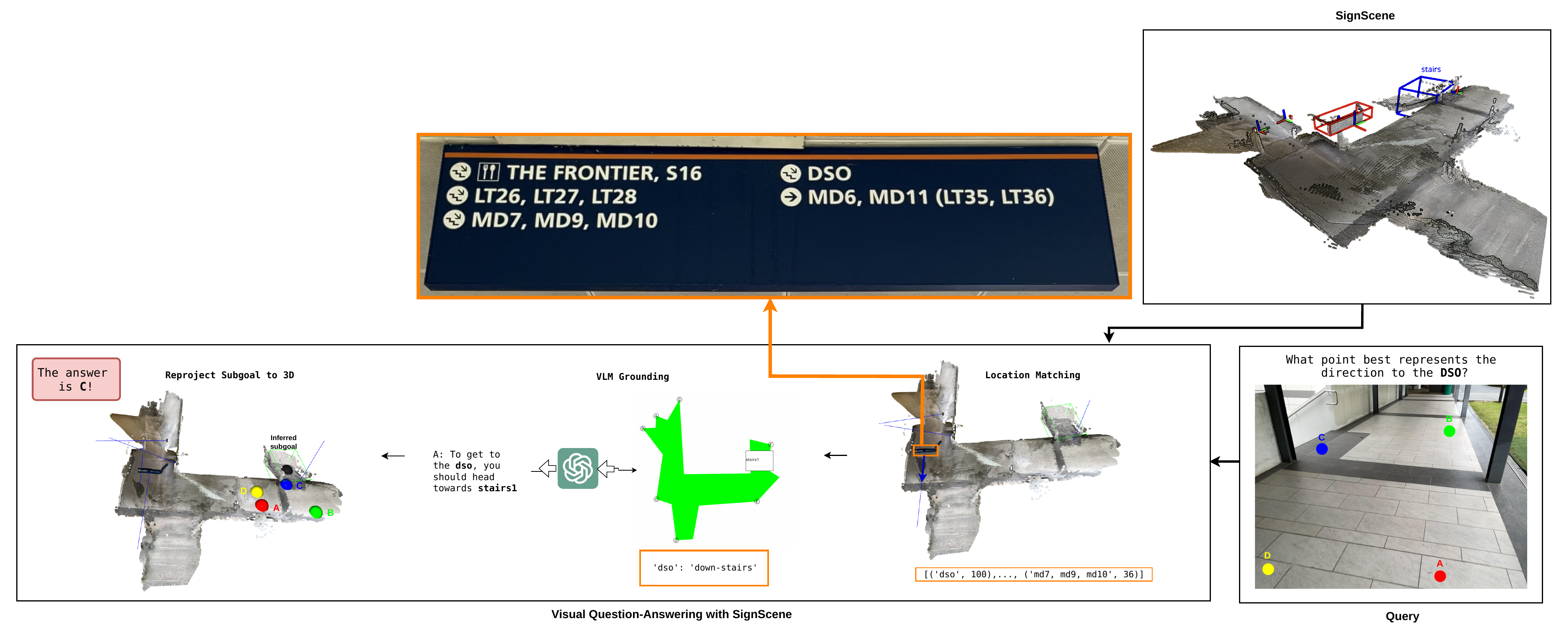}
  \caption{\textbf{Walking path connecting different campus buildings}. \method{} selects the sign in the \method{} container~(highlighted in orange) as being relevant to the query of \textbf{DSO}. It grounds the queried location to a navigation subgoal in the \method{} render. This subgoal is closest to point C in the query, resulting in the answer ``C''.}
  \label{fig:S64} 
\end{figure*}

\begin{figure*}[h]
  \centering
  \includegraphics[width=0.95\linewidth]{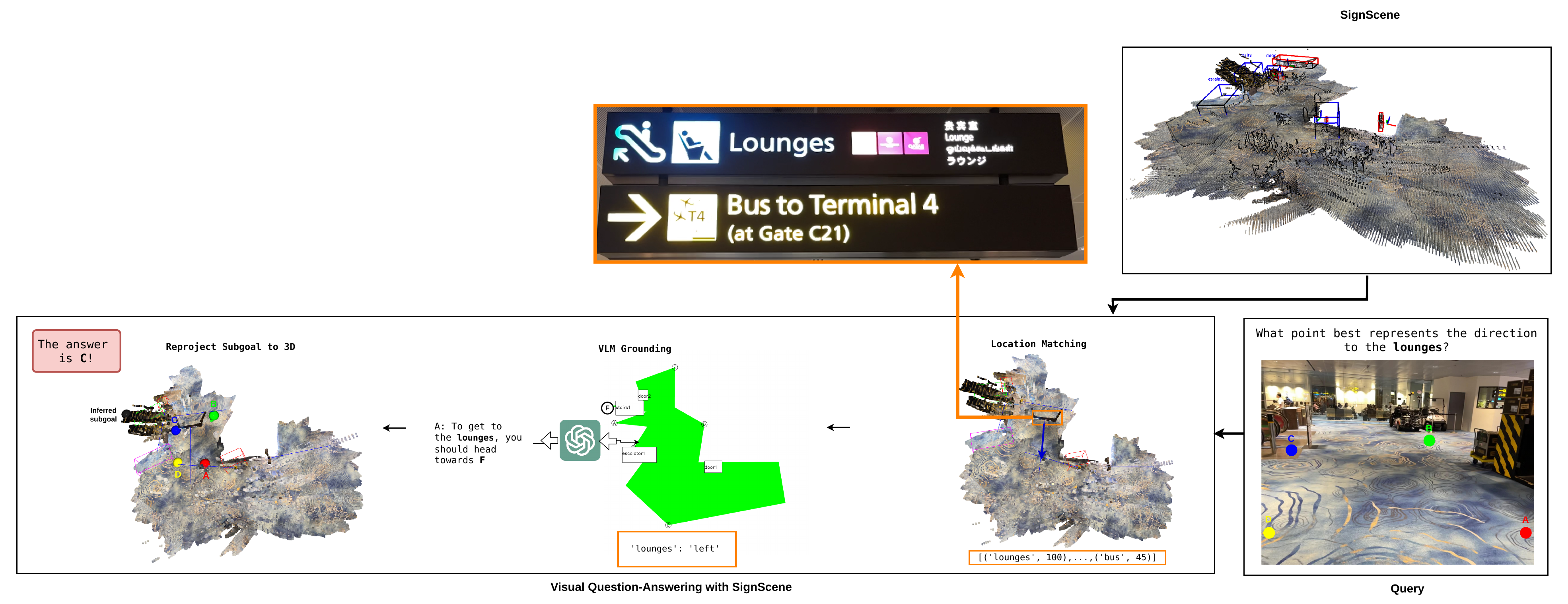}
  \caption{\textbf{Airport terminal}. \method{} selects the sign in the \method{} container~(highlighted in orange) as being relevant to the query of \textbf{lounges}. It grounds the queried location to a navigation subgoal in the \method{} render. This subgoal is closest to point C in the query, resulting in the answer ``C''.}
  \label{fig:S86} 
\end{figure*}

\begin{figure*}[h]
  \centering
  \includegraphics[width=0.95\linewidth]{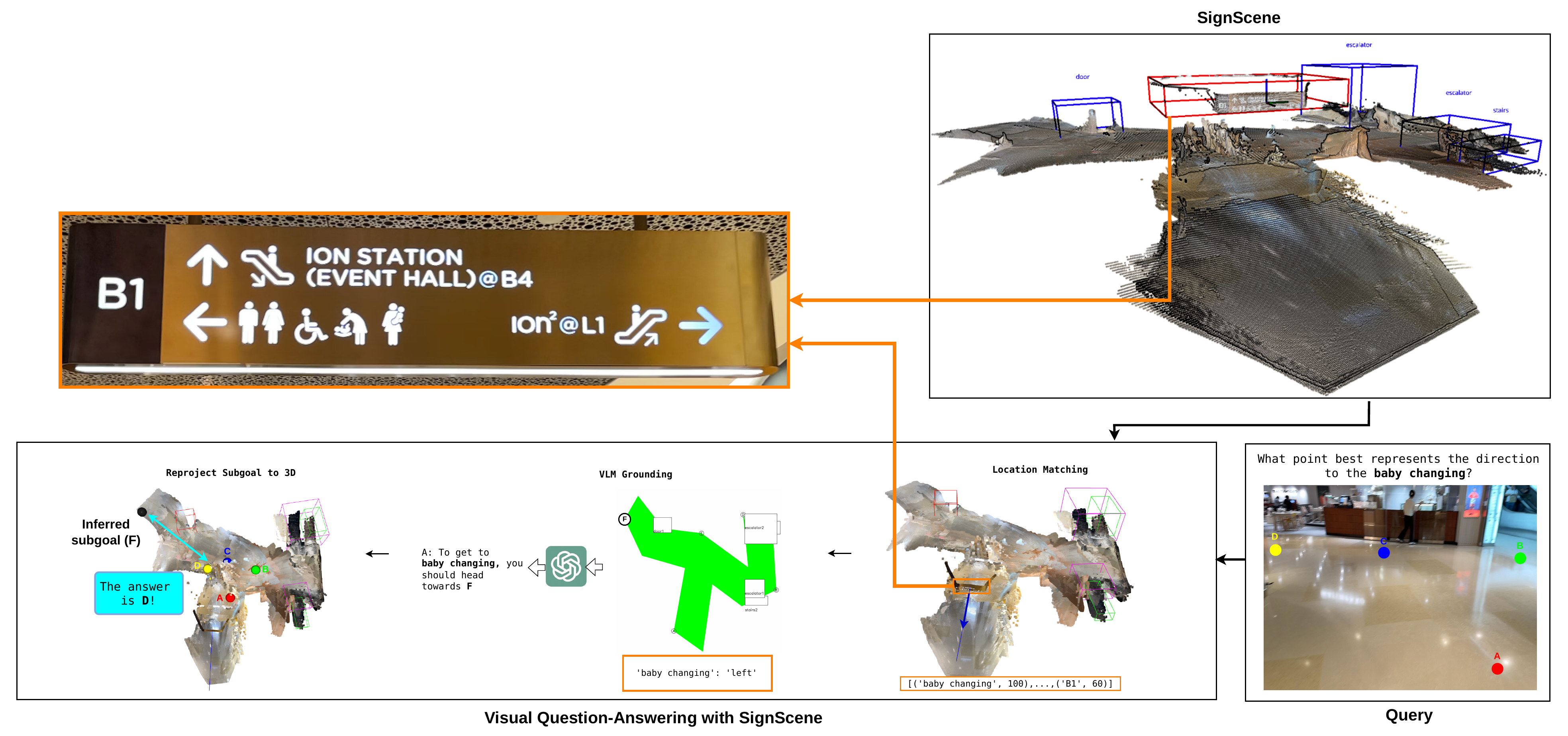}
  \caption{\textbf{Shopping mall}.  \method{} selects the sign in the \method{} container~(highlighted in orange) as being relevant to the query of \textbf{baby changing}. It grounds the queried location to a navigation subgoal in the \method{} render. This subgoal is closest to point D in the query, resulting in the answer ``D''.}
  \label{fig:S36} 
\end{figure*} 

\clearpage

\section*{Appendix B: Prompts}
\begin{figure}[]
  \centering
  \includegraphics[width=0.5\linewidth]{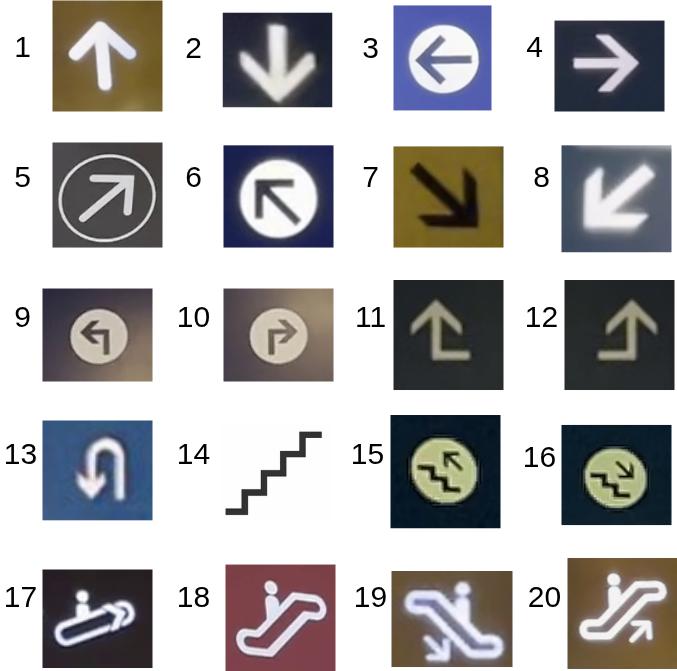}
  \caption{\textbf{Symbol Dictionary.}}
  \label{fig:directionsymbols} 
\end{figure}

For sign parsing, we apply in-context learning and first prompt~(\autoref{listing:incontext_prompt}) the VLM with the symbol dictionary~(\figref{fig:directionsymbols}), that includes the visual prototypes and a JSON with corresponding labels.
Then we provide the VLM with the sign image we would like to parse, and a detailed prompt~(\autoref{listing:parse_prompt}) specifying how the output should be parsed. 
For the visual grounding queries, the VLM is prompted with the sign-centric \method{} and description of how to select an answer~(\autoref{listing:ground_prompt}). 

\begin{lstlisting}[language=json, frame=single, caption={Labels for symbol dictionary}]
{
    "1" : "forward",
    "2" : "backwards",
    "3" : "left",
    "4" : "right",
    "5" : "forward-right",
    "6" : "forward-left",
    "7" : "backwards-right",
    "8" : "backwards-left",
    "9" : "forward-then-left",
    "10" : "forward-then-right",
    "11" : "left-then-forward",
    "12" : "right-then-forward",
    "13" : "u-turn",
    "14" : "stairs",
    "15" : "up-stairs",
    "16" : "down-stairs",
    "17" : "travelator",
    "18" : "escalator",
    "19" : "down-escalator",
    "20" : "up-escalator"
}
\end{lstlisting}

\begin{llmpromptbox}[specstyle, label=listing:incontext_prompt]{In-context learning prompt}

\begin{lstlisting}[style=Prompt]
The symbols in the image are associated to their semantic meaning through their index. This following dictionary {symbolDictionary} includes their semantic meaning.
\end{lstlisting}
\end{llmpromptbox}

\begin{llmpromptbox}[specstyle, label=listing:parse_prompt]{Sign parsing prompt}
\begin{lstlisting}[style=Prompt]
Break down the directions sign you see in the image. If the text is the illegible/unreadable, do not consider it. 
Return your answer in the form 
{
    'forward': [content], 
    'backwards': [content],
    'left': [content], 
    'right': [content],
    'forward-left': [content], 
    'forward-right': [content], 
    'backward-right': [content], 
    'backward-left': [content], 
    'forward-then-left': [content], 
    'forward-then-right': [content], 
    'left-then-forward': [content], 
    'right-then-forward': [content], 
    'up-stairs' : [content], 
    'down-stairs' : [content], 
    'down-escalator': [content], 
    'up-escalator': [content],
    'locational': [content]
}.                           
Return only the dict (no comments or formatting).
\end{lstlisting}
\end{llmpromptbox}

\begin{llmpromptbox}[specstyle, label=listing:ground_prompt]{Sign grounding prompt with SignScene}
\begin{lstlisting}[style=Prompt]
Select a letter in a circle or a object name in a bounding box in this image that is potentially closest to {location}, given the following list that consist of direction and places: {parsing}. Remember that lifts, stairs and escalators allow you do go up and down. Reason carefully about the spatial layout and the affordance of the objects. Return the answer based on the example format: [A] / [Door].
\end{lstlisting}
\end{llmpromptbox}

\begin{figure*}[t]
  \centering
  \includegraphics[width=0.95\linewidth]{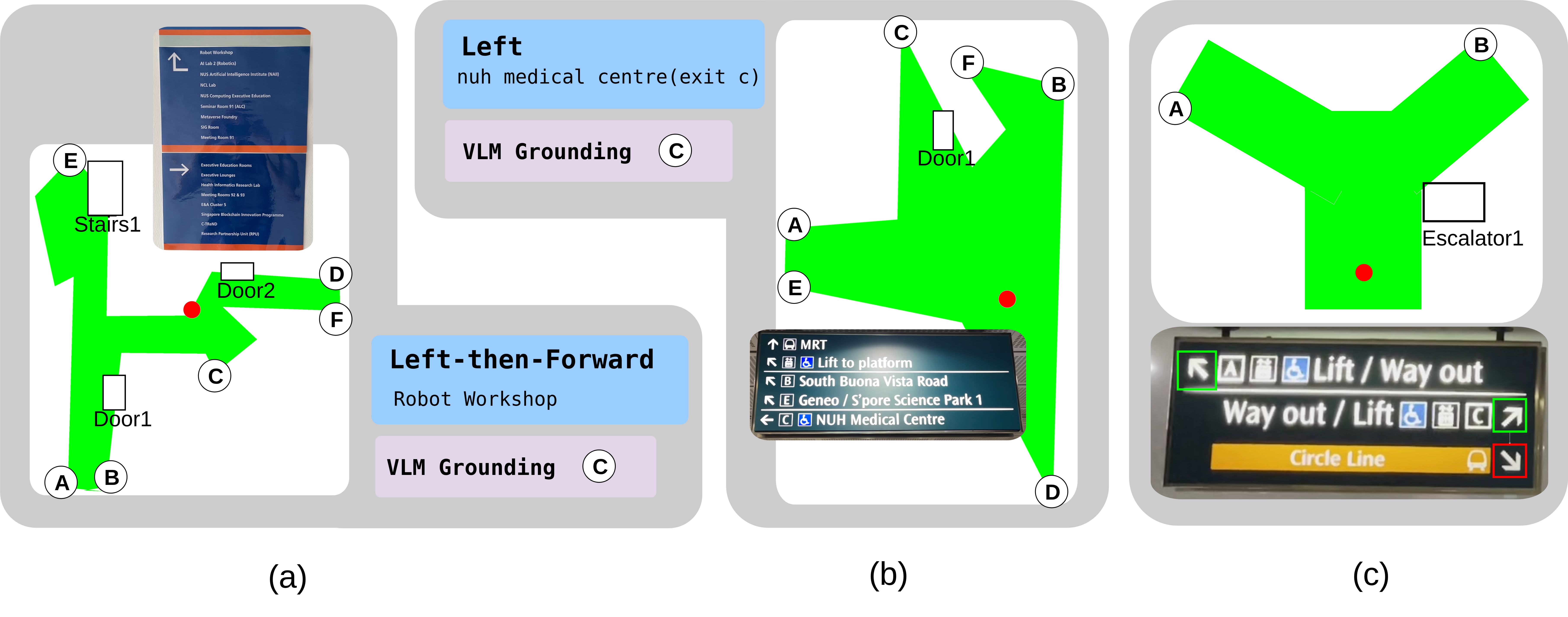}
  \caption{\textbf{Failure Analysis.} Figures (a) and (b) address \textbf{reasoning failure.} Despite accurate map construction and correct parsing, the VLM reasoning step can still fail. The red dot indicates the location of the sign. In (a) the VLM fails to ground correctly the compound instruction "left-then-forward". In (b) the VLM's choice of boundary point C is influenced by "exit c" in the parsed location.
  Figures (c) addresses \textbf{parsing in context.} All arrows are diagonal but the grounded navigational instructions is different. The green-annotated arrows refer to going forward-left and forward-right towards frontiers A and B, respectively. However, the red-annotated arrow refers to the down-moving escalator to the right, rather than walking backwards-right. } 
  \label{fig:failanalysis} 
\end{figure*}

\section*{Appendix C: Failure Analysis}

We identified 3 main failure modes for \method{}:
\begin{itemize}

    \item \textit{Object detection performance}. While open-set object detectors exhibit powerful generalization, they may still have poor precision and recall. We observed high rates of false positive and false negatives when detecting navigational structures like ``escalators'' with GroundingDINO. These rates varied across scene types, making it challenging to determine a single threshold for detection. Failures thus include both wrongly added objects or missing objects from the map. These inaccuracies cause difficulties in reasoning and degrade sign grounding performance.
    \item \textit{Parsing signs}. Compound instructions and those requiring additional scene context remain challenging for our sign understanding. For example (\figref{fig:motivation}), reaching ``ION station'' requires both heading straight and taking an escalator down to B4. The VLM fails to correctly decompose the instruction, interpreting the escalator as a location instead of a part of a compound instruction. Separately, in \figref{fig:failanalysis}(c), diagonal arrows encode different semantics that require scene context to decode: arrows in the green boxes point to different directions to take in the horizontal plane, whereas the arrow in the red box signals making a level change by taking a downward escalator on the right. These ambiguities lead to parsing errors that propagate and degrade downstream sign grounding.
    \item \textit{Reasoning}. VLMs' reasoning has several failure modes, even when given a correctly parsed sign and \method{}. VLMs often fail to ground more complex navigational instructions, such as "left-then-forward", as shown in \figref{fig:failanalysis}(a). Additionally, certain terms contained in the correctly parsed locations can misdirect the VLM: in \figref{fig:failanalysis}(b) the VLM appears to consistently select boundary point C as the answer due to the include of ``exit c'' as part of the location phrase ``medical centre (exit c)''. In doing so, it completely disregards the associated instruction $\leftarrow$.

\end{itemize}

\end{document}

%% file: macros.tex
\usepackage{xspace}
\usepackage[breakable]{tcolorbox}
\usepackage{listings}

\newcommand{\cmark}{\ding{51}}%
\newcommand{\xmark}{\ding{55}}%

\newcommand{\method}{\textbf{SignScene}}

\newcommand*{\signloc}[1][]{\mathcal{L}^{#1}}
\newcommand*{\signdir}[1][]{\mathcal{I}^{#1}}

\newcommand*{\dirregion}[1][]{\mathcal{R}^{#1}}
\newcommand{\navcues}{\mathcal{C}}
\newcommand{\navsign}{\mathcal{S}}
\newcommand{\obs}{I_t}

\newcommand{\odom}{u_t}
\newcommand{\depth}{D_t}
\newcommand{\depthseg}{s_{\text{pcd}}^i}
\newcommand{\signmask}{\mathcal{M}_{\text{sign}}}
\newcommand{\pathmask}{\mathcal{M}_{\text{path}}}
\newcommand{\mapfull}{M_{\text{3D}}}
\newcommand{\mapsimple}{M_{\text{2D}}}

\def\secref#1{Sec.~\ref{#1}}
\def\figref#1{Fig.~\ref{#1}}
\def\tabref#1{Tab.~\ref{#1}}
\def\eqref#1{Eq.~(\ref{#1})}

\makeatletter
\usepackage{xspace}
\DeclareRobustCommand\onedot{\futurelet\@let@token\@onedot}
\def\@onedot{\ifx\@let@token.\else.\null\fi\xspace}
\def\eg{e.g\onedot} 
\def\ie{i.e\onedot}

\def\etal{{et al}\onedot}
\def\etalcite#1{\etal~\cite{#1}}

\makeatother

\def\weblink{\urllink[pre = \bgroup\bf, post = \egroup]}

\lstdefinestyle{Prompt}{
  basicstyle        = \small\ttfamily,
  showstringspaces  = false,
  breaklines        = true,
  breakindent       = 0pt,
  rulecolor         = \color{black},
}

\tcbset{
  specstyle/.style={
    colframe=darkgray, 
    coltitle=white, 
    fonttitle=\bfseries, 
    boxrule=0.5mm, 
    rounded corners,
    sharp corners=south, 
    width=\linewidth, 
    boxsep=4pt, 
    left=8pt, 
    right=8pt, 
    top=2pt, 
    bottom=2pt, 
    before=\bigskip, 
    after=\bigskip, 
    breakable,
  }
}

\newtcolorbox[auto counter]{llmpromptbox}[2][]{%
    title=VLM Prompt ~\thetcbcounter: #2, #1}

\expandafter\def\csname tcb@cnt@llmpromptboxautorefname\endcsname{VLM Prompt}